\documentclass{article}

\usepackage{amsmath,amssymb,mathtools,bm}
\usepackage{graphicx}
\usepackage{booktabs}
\usepackage{algorithm}
\usepackage{algorithmic}
\usepackage[hidelinks]{hyperref}

\title{EPIG-Tree: Compute-Optimal Branching for Gradient-Efficient Reinforcement Learning}
\author{Nikita Khomich \and Leopold Hermansson \and Ido Hakimi}
\date{}

\begin{document}
\maketitle

\begin{abstract}
Reward based reinforcement learning for language models, exemplified by Group Relative Policy Optimization (GRPO), collapses an entire stochastic trajectory into a single scalar reward. This is clean and scalable, but it explores and allocates reward inefficiently: a trajectory may contain many causal decisions, recovery attempts, and environment-randomness events, yet every token or action inherits one trajectory-level advantage. We study tree-based rollout construction as a compute-allocation problem for policy-gradient estimation. Our central claim is that tree branches should not be placed where the policy is merely uncertain; they should be placed where an additional branch most reduces uncertainty about the policy gradient accounting for the expected compute cost. From a law-of-total-variance decomposition of the local policy-gradient random variable, we derive two allocation laws: new branches reduce decision uncertainty, while repeated suffix rollouts reduce continuation uncertainty. The resulting EPIG-Tree score allocates branches using the already computed rollouts. It estimates occupancy and score weighted value uncertainty, along with a suffix law
\[
n_e \propto \frac{w_e\,\|\nabla_\theta \log \pi(a_e\mid h_e)\|\,\sigma_e}{\sqrt{c_e}}.
\]
Empirically, we show that EPIG reduces gradient MSE in cloned-state control, winning in all nine dense continuous-control environments of a thirteen-environment sweep and recovering the reference gradient direction near-perfectly, and it improves frozen-LLM gradient calibration relative to simpler entropy branching. In online single-turn math, tree-local credit beats flat GRPO, while branch placement is secondary to token-level credit assignment. In online multi-turn Wordle, EPIG attains the highest final win rate (0.850), overtaking flat GRPO, which saturates early at 0.790, and entropy branching as training proceeds, confirming that the gradient-estimation advantage transfers to a stateful, large-action setting.
\end{abstract}

\section{Introduction}
The recent success of reinforcement learning with verifiable rewards has made on-policy algorithms such as GRPO central to LLM reasoning training~\cite{deepseekmath}. GRPO samples a group of responses for the same prompt, normalizes terminal rewards within the group, and applies a PPO-style clipped objective without training a separate critic.

GRPO nonetheless has three structural failure modes in long-horizon or multi-turn settings.

\textbf{Poor recovery credit assignment.} Suppose that a trajectory makes a mistake while performing correct actions and fails. Reward based GRPO assigns the whole trajectory a negative group-relative reward, so the correct actions are also punished. Conversely, if a trajectory succeeds despite multiple bad actions, the bad actions are reinforced. Recovery behavior is thus not a special case of long-horizon credit assignment; it is a direct consequence of assigning one scalar to a whole path.

\textbf{Inefficient handling of nondeterminism.} In a stochastic environment, a terminal reward is a noisy observation of $Q(h,a)=\mathbb{E}[R\mid h,a]$. A single trajectory conflates policy choice with environment randomness. If the environment randomly sabotages a good action, GRPO punishes the action; if it rescues a bad action, GRPO reinforces it. Estimating conditional continuation values requires repeated counterfactual suffixes from the same state.

\textbf{Poor scaling with rollout count.} Increasing the number of independent full rollouts explores more, yet many rollouts differ only in low-importance tokens, formatting, or environment noise. These do not provide any signal for the model to learn with.

Tree rollouts are a natural remedy. Instead of sampling $K$ independent full trajectories, we reuse prefixes, branch at intermediate states, and estimate descendant values. This yields local, process-like advantages without training a separate process reward model. Recent LLM tree methods follow this direction: TreeRL's EPTree forks from high-entropy intermediate tokens~\cite{treerl}; TreePO uses local uncertainty and segment-level tree modeling to amortize common prefixes~\cite{treepo}; and related tree-structured rollout methods derive relative advantages in multi-turn agent tasks. The unresolved question is: \emph{where should the tree branch?}

A common answer is entropy. Entropy is useful: if a policy is deterministic at a prefix, its branches coincide. But entropy alone is not the quantity that the policy-gradient estimator cares about. A high-entropy choice among reward-equivalent phrasings is not worth branching; a moderate-entropy choice between high- and low-value actions can be crucial. The organizing principle of this paper is therefore:

\begin{quote}
\textbf{A tree is an experimental apparatus for estimating the policy gradient.}

Optimal branching allocates compute where an additional branch most reduces gradient-estimator error per unit cost---not where the policy is merely uncertain.
\end{quote}

Concretely: GRPO collapses a full trajectory into one scalar advantage, whereas a tree estimates local conditional values; entropy measures only whether branches can differ, while EPIG measures whether they matter for the gradient.

\textbf{Contributions.}
\begin{enumerate}
\item We derive a law-of-total-variance decomposition showing that tree construction has two distinct jobs: new branches reduce decision uncertainty, while suffix resampling reduces continuation uncertainty.
\item We prove a compute-optimal suffix allocation law
\[
n_e^* \propto \frac{w_e\,\|\nabla \log \pi(a_e\mid h_e)\|\,\sigma_e}{\sqrt{c_e}}
\]
and a marginal branching law based on occupancy and score weighted value uncertainty.
\item We introduce EPIG-Tree, an algorithm that uses entropy as a proposal mechanism but allocates branches by expected predictive information gain about the gradient.
\item We validate the theory across cloned-state continuous control, frozen LLM gradient calibration, single-turn LLM math, and online multi-turn Wordle, establishing both where EPIG wins and the boundary conditions under which it ties cheaper baselines.
\end{enumerate}

\section{Related work}
\textbf{Policy gradients and PPO.} Our derivation begins from the standard policy-gradient theorem~\cite{sutton1999} and PPO-style clipped optimization~\cite{ppo}. The novelty is not a new policy-gradient identity; it is the use of the identity to formulate tree construction as optimal sampling for the gradient.

\textbf{GRPO and RLVR.} DeepSeekMath introduced GRPO as a critic-free PPO variant for mathematical reasoning~\cite{deepseekmath}, replacing a learned value baseline with group-relative normalization. Our critique is that the group-normalized terminal reward is a coarse estimator of local advantages in long-horizon settings.

\textbf{Tree search for LLM RL.} TreeRL proposes EPTree, an entropy-guided tree search that forks from high-uncertainty tokens and derives process supervision from descendant correctness~\cite{treerl}. TreePO similarly uses local uncertainty with segment-level tree modeling and prefix amortization~\cite{treepo}. These methods motivate our setting; we replace entropy/uncertainty heuristics with a gradient-estimation objective.

\textbf{Entropy in LLM RL.} The entropy-mechanism analysis argues that RL for reasoning models trades policy entropy for downstream performance, deriving entropy change from the covariance between logit updates and action probabilities/advantages~\cite{entropymechanism}. Complementary work finds that a minority of high-entropy tokens act as reasoning forks and carry much of the RLVR update signal~\cite{entropyminority}. We agree that entropy identifies branchable or update-sensitive points. Our result is narrower and sharper: entropy is not the objective; it is one input to a compute-allocation problem.

\textbf{Reward overoptimization.} Gao, Schulman, and Hilton study gold reward as a function of optimization distance $d=\sqrt{D_{\mathrm{KL}}(\pi\|\pi_0)}$, finding distinct scaling forms for best-of-$n$ and RL optimization and Goodhart-style degradation under proxy rewards~\cite{gao2023}. We include a Goodhart-corrected EPIG-Full variant for proxy-reward settings, while our strongest evidence concerns exact or cloned-state gradient estimation.

\section{From outcome GRPO to local gradient experiments}
Consider a history or prefix $h$, an action $a\sim\pi_\theta(\cdot\mid h)$, and a utility
\begin{equation}
Y = U\bigl(G(h)+R_{\mathrm{future}}\bigr),
\end{equation}
where $G(h)$ is reward already accumulated and $U$ is the training utility (linear return, success/failure, preference utility, verifier score, or task-specific reward). Define
\begin{equation}
Q_h(a)=\mathbb{E}[Y\mid h,a],\qquad
\sigma_h^2(a)=\mathrm{Var}(Y\mid h,a),\qquad
\psi_h(a)=\nabla_\theta\log\pi_\theta(a\mid h).
\end{equation}
The local contribution to the policy gradient is
\begin{equation}
g_h=\mu_h\,\mathbb{E}_{a\sim\pi_h}\bigl[\psi_h(a)Q_h(a)\bigr],
\end{equation}
where $\mu_h$ is the occupancy or training weight of the prefix. For an LLM token prefix, $\mu_h$ captures how frequently the current policy reaches that prefix or how much training mass we assign to it; for a cloned Gym/MuJoCo state, $\mu_h$ may be uniform over sampled states.

Outcome-only GRPO estimates many local terms with a single trajectory scalar. With $K$ sampled completions $\tau_i$ for a prompt, the group advantage is
\begin{equation}
\widehat A_i^{\mathrm{GRPO}} = \frac{R_i-\bar R}{s_R+\varepsilon},
\end{equation}
and every action on trajectory $i$ receives this same sign and scale. A tree estimator instead estimates $Q_h(a)$ by descendant rollouts from a shared prefix:
\begin{equation}
\widehat Q_h(a)=\frac{1}{n_{h,a}}\sum_{j=1}^{n_{h,a}}Y_{h,a,j},\qquad
\widehat A(h,a)=\widehat Q_h(a)-\widehat V(h).
\end{equation}
The mathematical question is now: given a compute budget, where should we spend the next suffix rollout or branch?

\section{The variance decomposition that determines the tree}
Let the random vector for a one-sample local gradient contribution be
\begin{equation}
Z_h=\mu_h\psi_h(A)Y,\qquad A\sim\pi_h.
\end{equation}
Its total variance decomposes as follows.

\textbf{Proposition 1 (Decision and continuation uncertainty).} For fixed prefix $h$,
\begin{align}
\mathrm{Var}(Z_h\mid h)
&=\mu_h^2\,\mathrm{Var}_{A\sim\pi_h}\bigl[\psi_h(A)Q_h(A)\bigr]\\
&\quad+\mu_h^2\,\mathbb{E}_{A\sim\pi_h}\bigl[\psi_h(A)\psi_h(A)^\top\sigma_h^2(A)\bigr].
\end{align}
Taking the trace gives a scalar gradient-MSE proxy.

\textit{Proof.} Apply the law of total variance,
$\mathrm{Var}(Z_h\mid h)=\mathrm{Var}(\mathbb{E}[Z_h\mid A,h]\mid h)+\mathbb{E}[\mathrm{Var}(Z_h\mid A,h)\mid h]$.
Since $\mathbb{E}[Z_h\mid A=a,h]=\mu_h\psi_h(a)Q_h(a)$ and $\mathrm{Var}(Y\mid h,a)=\sigma_h^2(a)$, the result follows. $\square$

This decomposition is the core of the paper. The first term is \emph{decision uncertainty}: uncertainty over which action branch matters. The second is \emph{continuation uncertainty}: noise in the suffix after a branch action has already been selected. Tree construction therefore has two distinct compute actions:
\begin{itemize}
\item add a new branch from $h$ to reduce $\mathrm{Var}_a[\psi_h(a)Q_h(a)]$;
\item add suffix samples under an existing edge $(h,a)$ to reduce $\sigma_h^2(a)$.
\end{itemize}

\subsection{Why entropy is insufficient}
Entropy is $H_h=H(\pi_\theta(\cdot\mid h))$: it measures how many alternatives the policy can produce. But the variance decomposition contains no standalone entropy term. A high-entropy prefix with $Q_h(a)$ nearly constant has little gradient-relevant decision uncertainty; a moderate-entropy prefix with large $Q_h$ dispersion can dominate the gradient MSE. Entropy is therefore a useful proposal mechanism---it tells us where branches can differ. EPIG asks whether those differences matter.

\section{Optimal allocation laws}
\subsection{Suffix allocation}
Let edge $e=(h,a)$ have training weight $w_e$, score vector $\psi_e=\nabla\log\pi(a\mid h)$, continuation standard deviation $\sigma_e$, and suffix cost $c_e$. If $n_e$ independent suffixes are allocated under $e$, the trace-variance contribution is approximately $A_e/n_e$ with
\[
A_e=w_e^2\|\psi_e\|^2\sigma_e^2.
\]

\textbf{Theorem 1 (Cost-sensitive suffix allocation).} For fixed candidate edges and budget $B$, the solution of
\[
\min_{n_e>0}\sum_e\frac{A_e}{n_e}\qquad\text{s.t.}\qquad \sum_e c_en_e\le B
\]
is
\begin{equation}
n_e^*\propto\sqrt{\frac{A_e}{c_e}}=\frac{w_e\|\psi_e\|\sigma_e}{\sqrt{c_e}}.
\end{equation}

\textit{Proof.} The Lagrangian is $\mathcal L(n,\rho)=\sum_e A_e/n_e+\rho(\sum_e c_en_e-B)$. Stationarity gives $-A_e/n_e^2+\rho c_e=0$, hence $n_e=\sqrt{A_e/(\rho c_e)}$; the proportionality follows after normalizing to the budget. $\square$

The law is a Neyman-allocation analogue for policy gradients: sample more where gradient leverage and suffix noise are high and cost is low.

\subsection{Branch allocation}
Let $m_h$ be the number of action branches already sampled at node $h$, and define the decision-uncertainty coefficient
\[
B_h=\mu_h^2\,\mathrm{tr}\!\left(\mathrm{Var}_{a\sim\pi_h}[\psi_h(a)Q_h(a)]\right).
\]
With $m_h$ independent branch samples, the action Monte Carlo term scales as $B_h/m_h$, so the discrete gain from one more branch is
\begin{equation}
\Delta_h^{\mathrm{branch}}\approx \frac{B_h}{m_h(m_h+1)}-\lambda c_h,
\end{equation}
where $c_h$ is expected branch/suffix cost and $\lambda$ is compute price. In practice $Q_h$ and $\psi_h$ are estimated from pilot branches, and EPIG uses the empirical score
\begin{equation}
S_{\mathrm{EPIG}}(h)=
\frac{\widehat\mu_h^{\,2}\,\mathrm{tr}\,\widehat{\mathrm{Var}}_{a\in\mathcal C(h)}\!\left[\widehat\psi_h(a)\widehat Q_h(a)\right]}
{\widehat c_h(m_h+1)^2+\varepsilon}.
\end{equation}
We call the full $\psi$-weighted score EPIG-grad---the variant used in all experiments below---and the value-variance-only score $\widehat{\mathrm{Var}}_a[\widehat Q_h(a)]/\widehat c_h$ EPIG-Lite. EPIG-Lite is cheaper and suffices in dense continuous control but can fail where score norms differ or action probabilities are very uneven. Our frozen Wordle calibration confirms the $\psi$-weighting is active: EPIG-grad beats pure value variance.

\textbf{Corollary 1 (Entropy branching as a special case).} Entropy-guided branching coincides with EPIG only under a homogeneity assumption: across candidate nodes, $\mu_h$, $\|\psi_h\|$, value dispersion, suffix noise, and cost must be constant or monotone functions of entropy. When these quantities decouple, entropy selects branchable but low-information prefixes.

\subsection{Goodhart and entropy-pricing extensions}
When rewards are exact or verifiable, we set Goodhart sensitivity to zero. For proxy rewards, local tree expansion can overoptimize the proxy. Following the empirical scaling-law view of reward-model overoptimization~\cite{gao2023}, define local optimization distance $d_h=\sqrt{D_{\mathrm{KL}}(\pi_h^{\mathrm{tree}}\|\pi_h)}$. A Goodhart-corrected gain is
\begin{equation}
G_h(d)=\int_0^d \alpha_h e^{H_h(u)}\,du-\beta_h\Phi(d),
\end{equation}
with $\Phi(d)=d^2$ for best-of-branch-like selection and $\Phi(d)=d\log d$ for RL-like updates. If $\kappa_h(d)=-H_h'(d)$ is entropy drain, the marginal equilibrium condition is
\begin{equation}
\frac{B_h}{m_h^2}+\mu_h\left[\alpha_h e^{H_h(d_h)}-\beta_h\Phi'(d_h)\right]d_h'(m_h)
=\lambda C_h'(m_h)+\tau\kappa_h(d_h)d_h'(m_h).
\end{equation}
We treat this as EPIG-Full. Our strongest measured evidence concerns the first term, the branch score, and the suffix law; the Goodhart and entropy-drain terms are theoretically motivated and remain to be stress-tested at larger proxy-reward scale.

\section{Algorithm}
EPIG-Tree uses entropy to propose candidate branch points but value/gradient information to allocate the branch budget.

\begin{algorithm}[htbp]
\caption{EPIG-Tree update for one prompt/state batch}
\begin{algorithmic}[1]
\REQUIRE policy $\pi_\theta$; prompts/states; root count $K_0$; branch budget $B$; pilot size $m_0$; compute price $\lambda$
\STATE Sample $K_0$ on-policy root trajectories; store prefixes/states $h$, actions $a$, log-probs, rewards, and costs.
\STATE Propose a candidate set $\mathcal C$ of branch states: turn or reasoning-step boundaries / high-entropy token segments (LLMs), or sampled clone states (cloneable RL).
\FOR{each candidate node $h\in\mathcal C$}
\STATE Score $S_{\mathrm{EPIG}}(h)$ using the branch score above.
\ENDFOR
\WHILE{branch budget $B$ remains}
\STATE Add a branch at $\arg\max_h S_{\mathrm{EPIG}}(h)$ with positive marginal gain; refresh its score.
\ENDWHILE
\STATE Allocate suffix rollouts per edge by $n_e^*\propto w_e\|\psi_e\|\sigma_e/\sqrt{c_e}$.
\STATE Form tree values $\widehat V(h)$ from descendant leaves and edge advantages $\widehat A(h,a)=\widehat V(\mathrm{child})-\widehat V(h)$ (optionally mixed with a root-relative advantage).
\STATE Apply a PPO/GRPO-style clipped update with an action-token / branch-segment mask, so the advantage trains the tokens that caused the branch.
\RETURN updated policy $\pi_\theta$.
\end{algorithmic}
\end{algorithm}

\section{Experiments}
We organize the evidence into three layers. The first directly tests the allocation law under frozen policies and cloneable states; the second tests frozen LLM gradient calibration; the third tests online LLM RL, where optimization dynamics and token credit assignment interact with branch placement. All numbers are measured.

\subsection{Cloned-state control: direct validation of the allocation law}
\textbf{Setup.} We freeze a policy, sample states, clone each state, sample candidate actions, and estimate a high-budget reference gradient with 32 suffix rollouts per action. Each allocation method receives budget $B=8$ suffixes and estimates $\widehat g$; we report $\|\widehat g-g_{\mathrm{ref}}\|_2$ and cosine to the reference. We compare EPIG-grad against entropy, uniform, value variance, suffix-$\sqrt{\cdot}$, and oracle variants; the key comparison is EPIG-grad versus entropy and uniform.

\begin{figure}[!htbp]
\centering
\includegraphics[width=0.92\linewidth]{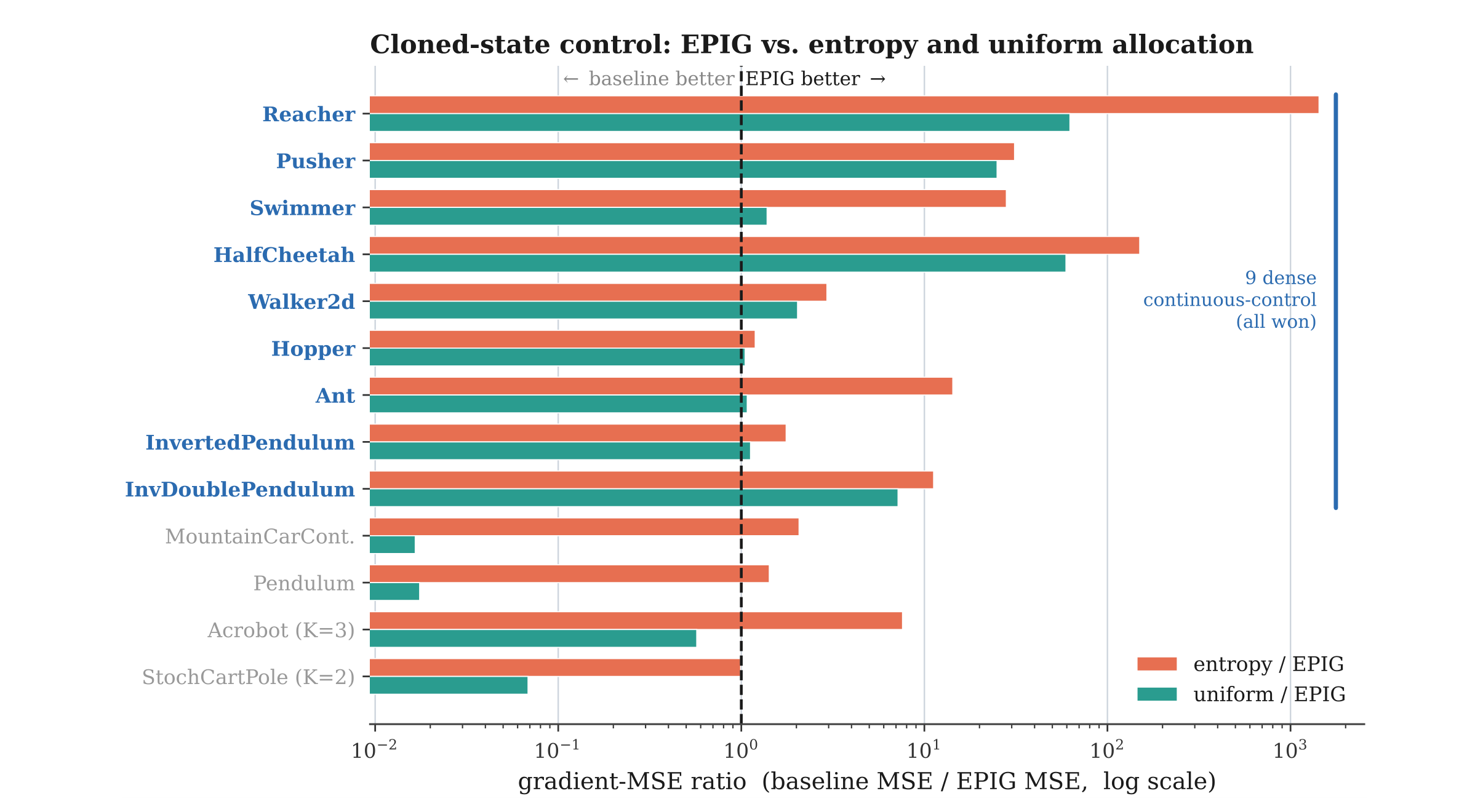}
\caption{Cloned-state control: entropy and uniform gradient-MSE divided by EPIG MSE (log scale). Values above 1 mean EPIG is better. EPIG wins on all nine dense continuous-control tasks and on the four sparse or tiny-action exceptions falls back toward the cheaper baselines.}
\label{fig:control-mse}
\end{figure}

\begin{figure}[!htbp]
\centering
\includegraphics[width=0.88\linewidth]{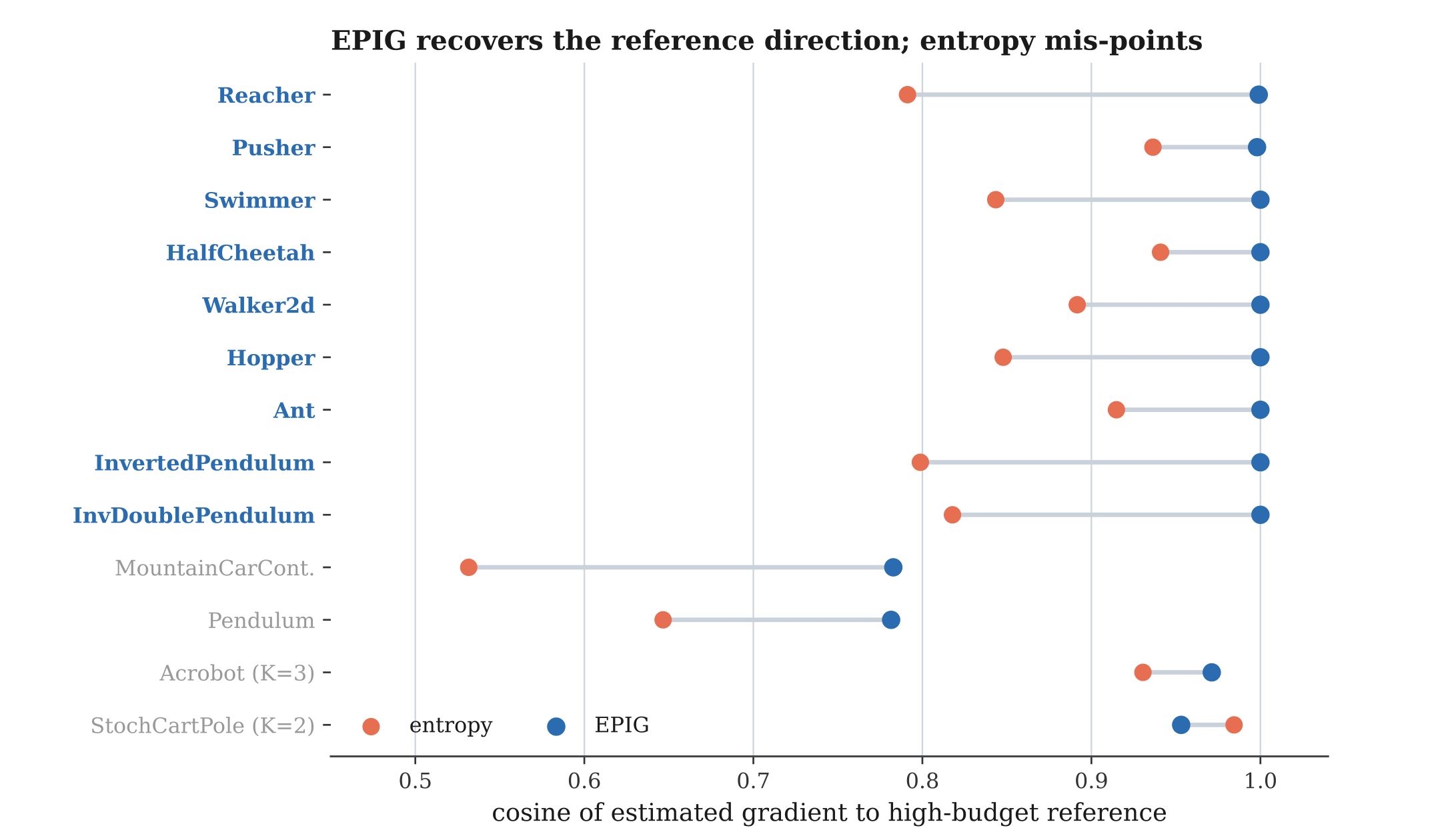}
\caption{Gradient cosine to the high-budget reference. On the nine dense continuous-control tasks EPIG recovers the reference direction near-perfectly (0.998--1.000) while entropy mis-points.}
\label{fig:control-cos}
\end{figure}

\textbf{Result.} EPIG-grad beats both entropy and uniform on gradient-MSE in 9/13 environments---exactly the nine dense continuous-control environments (Figure~\ref{fig:control-mse}). On these nine it recovers the reference gradient direction near-perfectly (cosine 0.998--1.000), while entropy lags and mis-points (cosine 0.79--0.94; Figure~\ref{fig:control-cos}). The four exceptions are informative rather than arbitrary. Tiny-action environments (Acrobot with $K=3$, a stochastic CartPole with $K=2$) leave little allocation problem---uniform already covers the relevant actions---and sparse or non-smooth reward regimes (MountainCarContinuous, Pendulum) make small-pilot value dispersion unreliable. EPIG is therefore not a universal replacement for exploration; it is a compute-allocation law for settings where local value differences can be estimated.

\subsection{Counterfactual branching helps when there is headroom}
\begin{figure}[!htbp]
\centering
\includegraphics[width=0.95\linewidth]{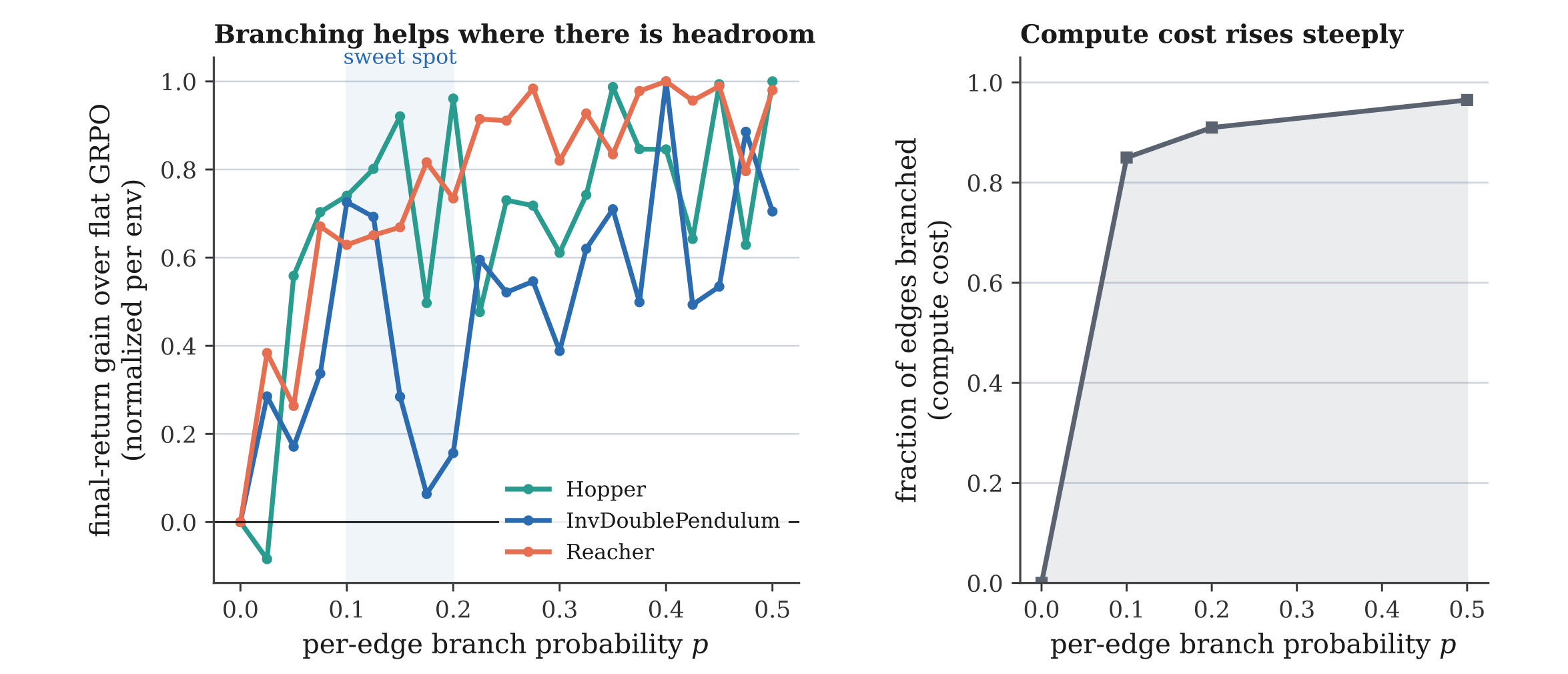}
\caption{Even unoptimized stochastic branching supplies useful counterfactual signal in control. Left: normalized final-return gain over flat GRPO versus branch probability $p$ for the three environments with headroom; the shaded band marks the $p\approx0.1$--$0.2$ sweet spot. Right: the branched-edge fraction (compute cost) rises steeply with $p$, so high $p$ buys little extra return.}
\label{fig:branching}
\end{figure}

Before selective EPIG allocation, a simpler question is whether counterfactual branching helps at all. In a seven-environment ablation, uniform per-edge stochastic branching---each edge forks with probability $p$, a fresh action is resampled, and all root and branch samples feed one group-normalized GRPO update---improves early learning wherever learning happens (Figure~\ref{fig:branching}). Relative to flat GRPO ($p=0$) under a capped update budget, InvertedDoublePendulum gains +52 final return, Hopper +16, and Reacher +5 at $p=0.5$, with the bulk of the gain captured by a moderate $p\approx0.1$--$0.2$. The mechanism is simply more counterfactual signal per update; the cost is proportionally more compute (branched-edge fraction $\to0.9+$ at high $p$), and gains diminish beyond the sweet spot. EPIG asks how to spend this branching budget more selectively.

\subsection{Frozen LLM gradient calibration}
\textbf{Setup.} We use Qwen3-8B on GSM8K and compare budgeted-tree PPO gradients against a 64-leaf high-budget reference over 64 prompts and 3 seeds. This is not an online training result; it asks whether the branch topology improves the gradient estimator.

\begin{figure}[!htbp]
\centering
\includegraphics[width=0.78\linewidth]{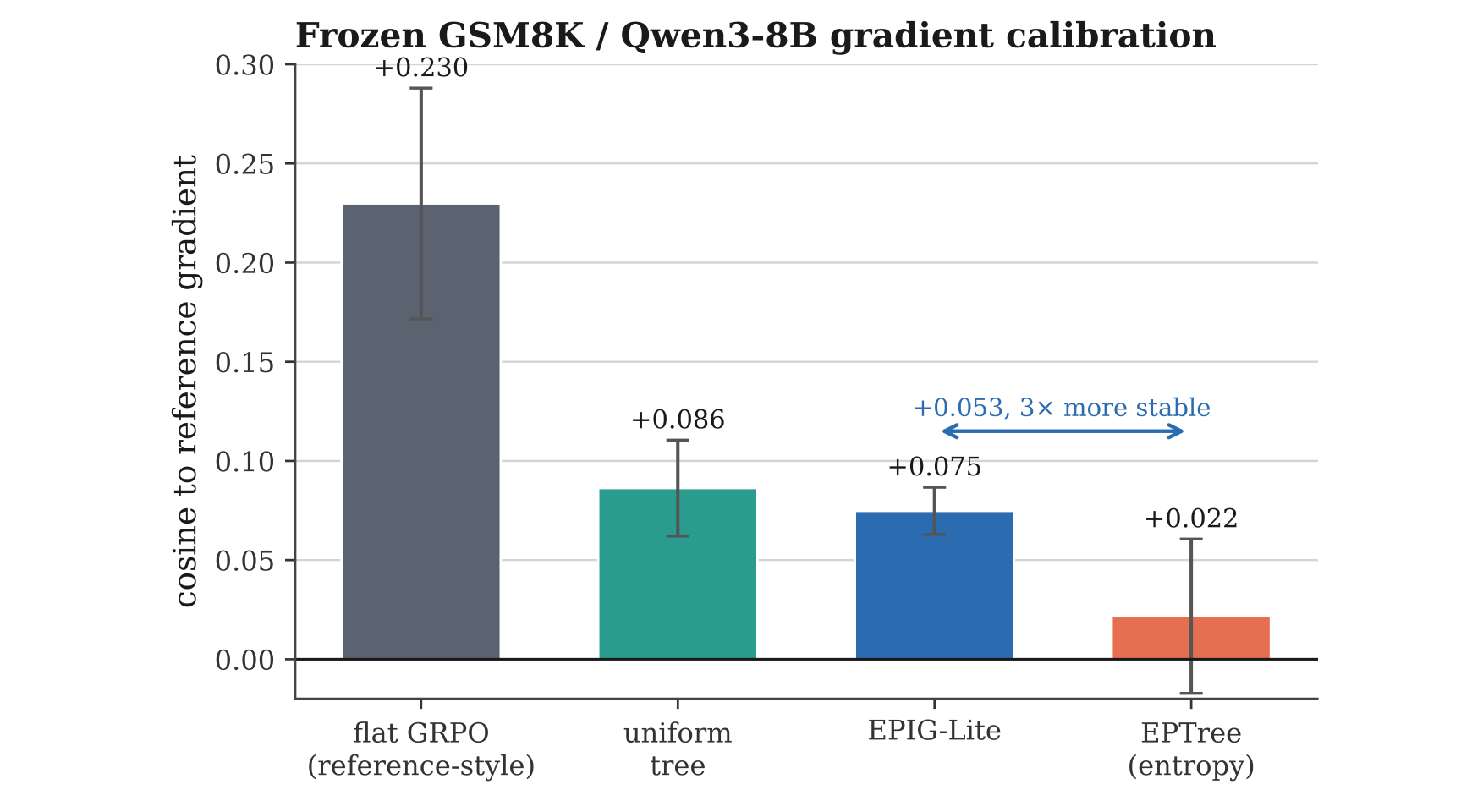}
\caption{Frozen GSM8K/Qwen3-8B gradient calibration. Among tree methods, EPIG is more stable and better aligned than EPTree-style entropy branching. Flat GRPO aligns best with the GRPO-style reference, highlighting that topology and advantage construction must be evaluated separately. Error bars are SEM over three seeds.}
\label{fig:frozen-gsm8k}
\end{figure}

\textbf{Result.} Among tree methods, EPIG-Lite is more stable and better aligned than entropy branching: cosine $0.0748\pm0.0119$ versus $0.0217\pm0.0389$ (entropy goes negative on one seed), a gap of +0.053 at one-third the standard deviation (Figure~\ref{fig:frozen-gsm8k}). Flat GRPO aligns best with the GRPO-style reference (0.2298), which we read as a warning that tree-local advantages and token masks are themselves consequential estimator choices---branch scoring is not the only thing a tree estimator must get right.

\subsection{Single-turn LLM math: tree credit helps, placement ties}
\begin{figure}[!htbp]
\centering
\includegraphics[width=0.9\linewidth]{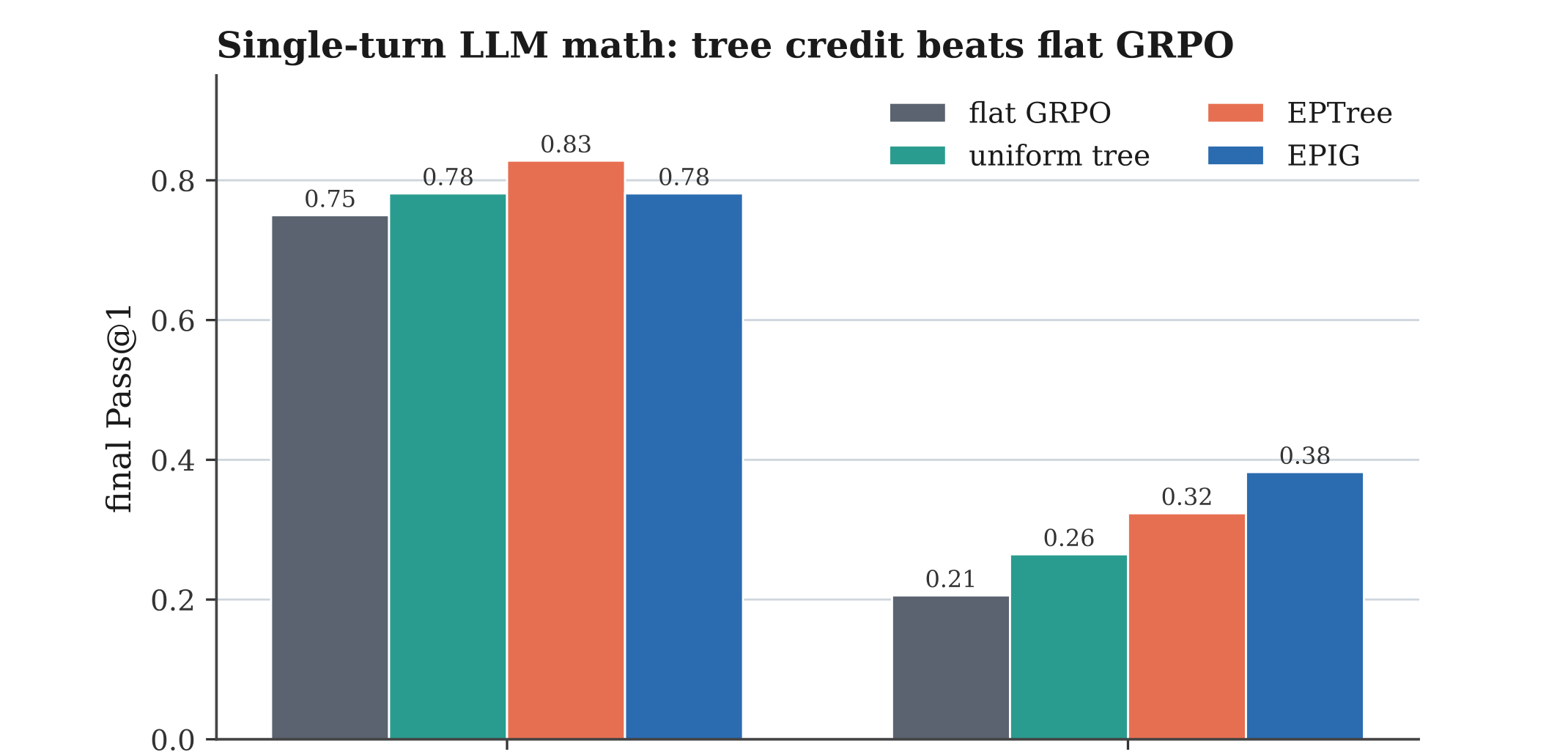}
\caption{Single-turn LLM math. Tree methods beat flat GRPO, especially on the harder MATH-small, but branch placement is within single-seed noise. This motivates stateful multi-turn benchmarks rather than more saturated single-turn math.}
\label{fig:math}
\end{figure}

On Qwen-class GSM8K-hard and MATH-small, tree-local training beats flat GRPO, but branch-placement variants fall within single-seed noise (Figure~\ref{fig:math}). EPIG wins MATH-small final Pass@1 (0.382 versus 0.206 for flat GRPO), while EPTree wins GSM8K-hard (0.828). We therefore do not claim that EPIG beats EPTree in single-turn math; the robust finding is that tree-local credit is useful and that token masking can dominate branch placement.

\begin{figure}[!htbp]
\centering
\includegraphics[width=0.9\linewidth]{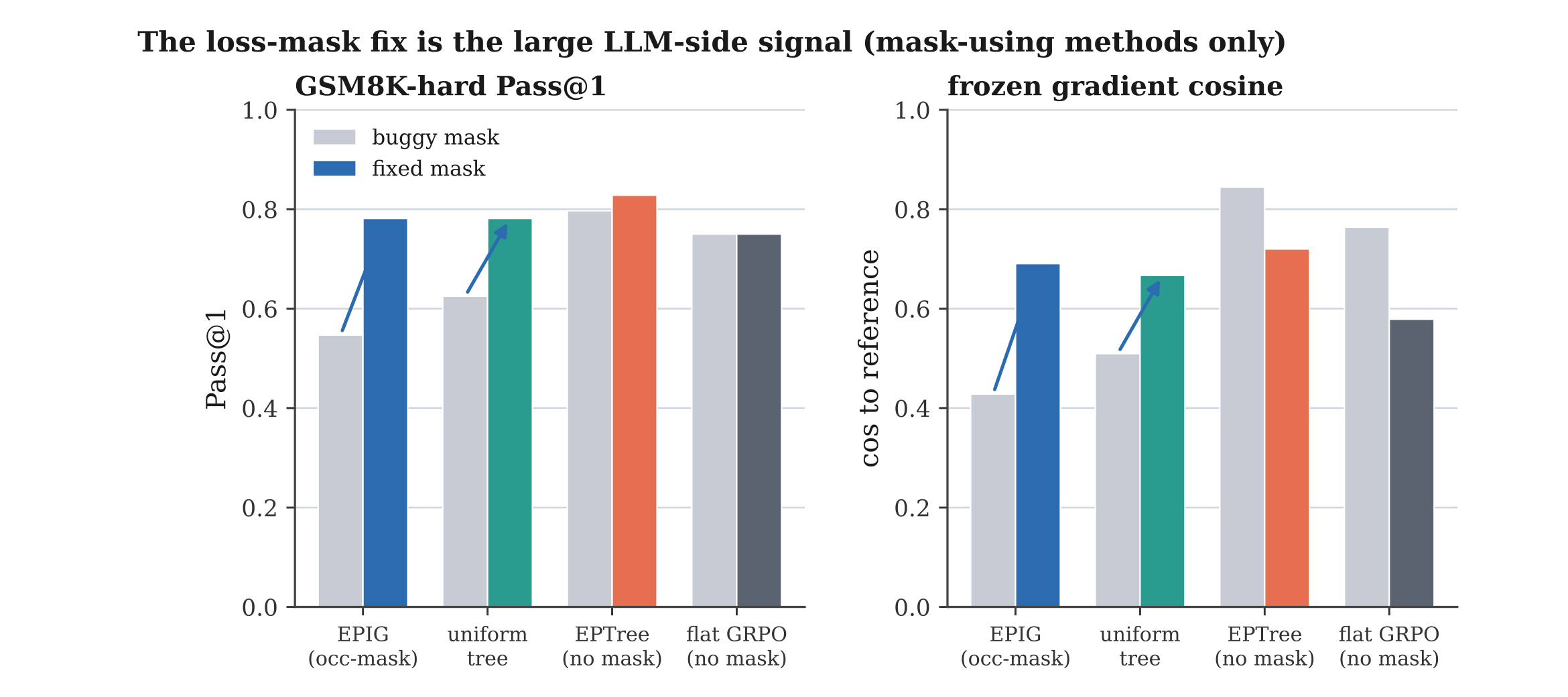}
\caption{The loss-mask fix is the large LLM-side signal. The two mask-using methods (EPIG, uniform tree) jump on both Pass@1 and frozen gradient cosine, while the non-mask methods (EPTree, flat GRPO) are essentially unchanged on Pass@1, isolating the mask as the cause.}
\label{fig:mask}
\end{figure}

The most consequential LLM engineering lesson was a loss-mask bug: an incorrect branch-segment mask trained only a short window and missed answer tokens in thinking-mode outputs. Fixing the mask moved EPIG on GSM8K-hard from 0.547 to 0.781 Pass@1 and frozen gradient cosine from 0.428 to 0.691, and moved uniform-tree from 0.625 to 0.781, while the two non-mask methods barely moved (Figure~\ref{fig:mask}). For LLM tree RL, topology and credit assignment are coupled: a good branch score is useless if its advantage is applied to the wrong tokens.

\subsection{Multi-turn Wordle: frozen calibration}
\begin{figure}[!htbp]
\centering
\includegraphics[width=0.86\linewidth]{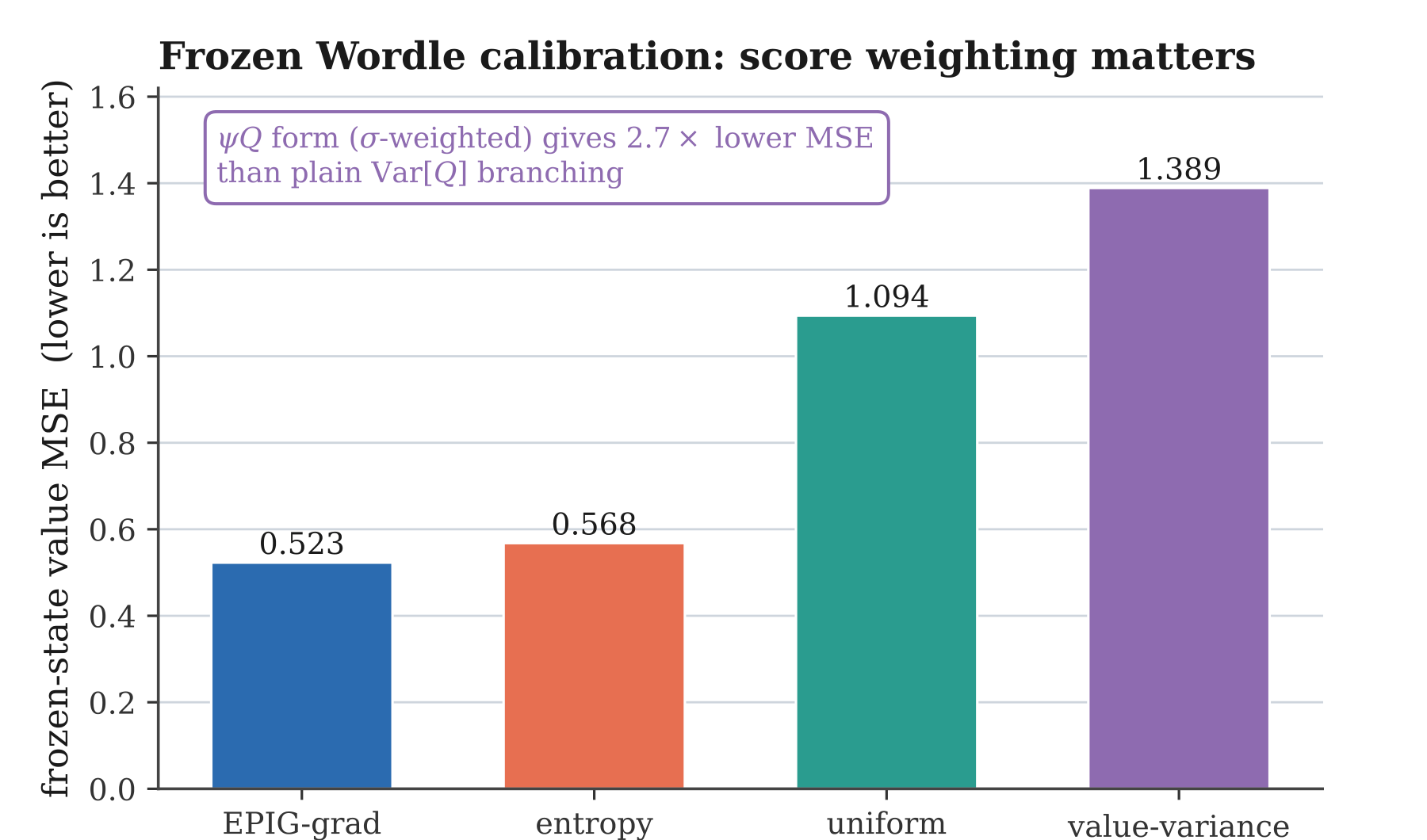}
\caption{Frozen Wordle state calibration. EPIG-grad has the lowest value MSE, and the gap to pure value variance shows that gradient/score weighting---not value dispersion alone---drives the advantage.}
\label{fig:wordle-frozen}
\end{figure}

Single-turn math distributes value over many reasoning tokens, making branch placement hard to isolate; multi-turn games provide clearer states and actions. On frozen Wordle states from a TextArena-style environment with a Wordle-tuned Qwen-family model, we estimate high-budget reference values and compare budgeted estimators. EPIG-grad achieves the lowest value MSE, 0.523, versus 0.568 for entropy and 1.094 for uniform (Figure~\ref{fig:wordle-frozen}). Crucially, pure value variance is worse (1.389): score weighting matters, supporting the theorem's $\psi Q$ form over a plain $\mathrm{Var}[Q]$ heuristic.

\subsection{Online multi-turn Wordle}
The decisive test is whether the dense-control and frozen-calibration advantage survives online, where the policy moves and the action space is large. We run online Wordle---a Wordle-tuned 1.7B model, four methods $\times$ two seeds, 300 policy updates---training each for the full update budget and logging win rate against environment interactions.

\begin{figure}[!htbp]
\centering
\includegraphics[width=0.92\linewidth]{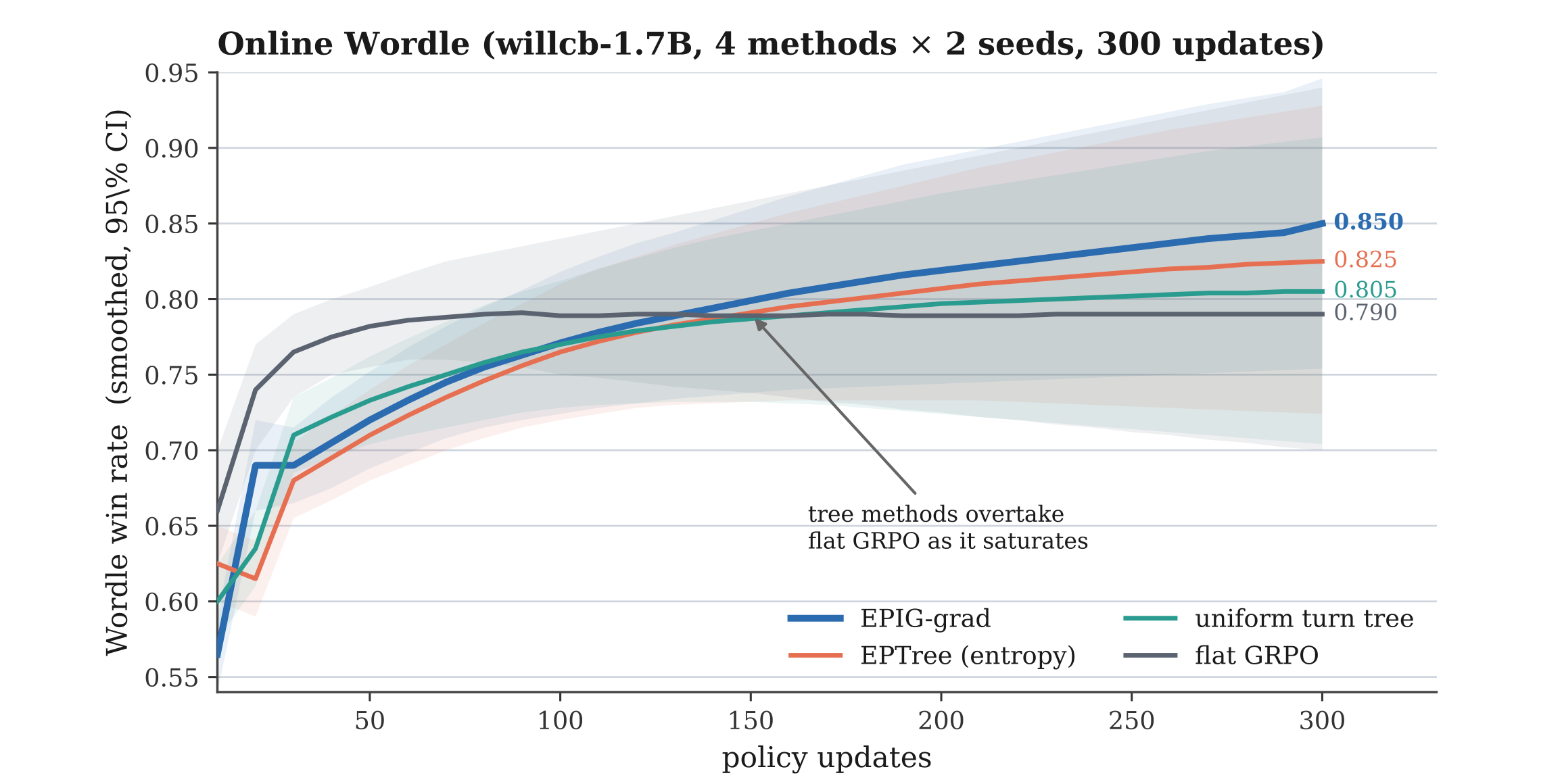}
\caption{Online Wordle win rate per policy update (smoothed, 95\% CI over two seeds). The tree methods start below flat GRPO but overtake it as it saturates near 0.790. Final win rates are annotated at right.}
\label{fig:wordle-online}
\end{figure}

The Wordle result is the clearest online evidence in the paper (Figure~\ref{fig:wordle-online}). All three tree methods start below flat GRPO in the first $\sim30$ updates---tree advantages are noisier early---but flat GRPO saturates near a win rate of 0.790 by update 50 and improves no further. The tree methods keep climbing and overtake it: EPIG-grad crosses flat GRPO around update 140 and separates over the second half of training, finishing at 0.850, compared with EPTree (0.825), uniform turn tree (0.805), and flat GRPO (0.790). The final EPIG margin over the strongest tree baseline is +0.025 win rate, and EPIG gains the most from extended training.

\section{Discussion}
The experiments support a nuanced conclusion. In theorem-aligned settings---especially cloned-state dense continuous control---EPIG allocation is strongly supported: it reconstructs the reference gradient direction where entropy mis-points. In frozen LLM diagnostics it improves tree-gradient alignment and Wordle value-MSE. In online single-turn LLM math, tree-local credit is helpful but branch placement is not the bottleneck; token loss masks and advantage construction dominate. In online multi-turn Wordle, where the policy moves and the action space is large, EPIG's allocation advantage reappears as a higher final plateau: it overtakes a fast-but-saturating flat GRPO and finishes ahead of every baseline.

\textbf{What the theorem does not say.} EPIG is not a universal exploration algorithm. If the action space is tiny, uniform coverage is already near-optimal. If rewards are sparse and pilot rollouts cannot reveal value differences, pure value-information allocation can be too exploitative. If branch advantages are applied to irrelevant tokens, topology does not matter. These are boundary conditions, not contradictions, and our control losses fall exactly where the boundary conditions predict.

\textbf{Why this matters for GRPO.} GRPO remains a powerful baseline, but in stochastic, multi-turn, or recovery-heavy environments its outcome-only estimator is inherently coarse. Trees expose local conditional values, and EPIG provides a principle for deciding which local counterfactuals are worth buying.

\section{Limitations}
Several LLM diagnostics use a single seed and should be read as preliminary rather than as ranked benchmarks; we mark single-seed comparisons as such throughout and do not overclaim sub-0.02 gaps. The Goodhart and entropy-drain terms are theoretically motivated and have not been stress-tested at large proxy-reward scale. Finally, EPIG depends on reasonably good pilot estimates of local value and score-weighted dispersion; sparse-reward environments may require additional exploration terms.

\section{Conclusion}
We proposed EPIG-Tree, a compute-optimal view of rollout-tree construction. The core derivation is simple: the variance of the local policy-gradient contribution splits into decision uncertainty and continuation uncertainty. New branches should reduce the former; suffix samples should reduce the latter. This yields explicit allocation laws and shows why entropy is an incomplete branching objective. The empirical picture matches the theory: EPIG wins cleanly in cloned-state control and frozen LLM diagnostics, ties cheaper baselines on near-saturated single-turn math, and recovers its advantage online in stateful, large-action multi-turn games. A tree is an experimental apparatus for the policy gradient, and EPIG is the design that spends its budget where the gradient is hardest to estimate.

\section*{Acknowledgments}
We thank our colleagues at Tzafon AI for infrastructure support and helpful discussions.

\newpage
\appendix
\section*{Detailed derivations}
\subsection*{Group-relative baselines and why local values help}
Ignoring standard-deviation normalization, the GRPO group baseline for sample $i$ is $R_i-\bar R$. If $\bar R$ includes $R_i$, the estimator is shrunk by a factor close to $(1-1/K)$ in the simplest independent setting; a leave-one-out baseline removes this shrinkage. The larger issue is variance and credit assignment: the same scalar multiplies every score term on the trajectory. Tree-local values approximate $Q(h,a)-V(h)$ at internal edges, reducing the chance that recovery actions inherit the sign of unrelated terminal events.

\subsection*{Connection to mutual information}
The action channel $A\to Y$ has utility information
\[
I(A;Y\mid h)=H(Y\mid h)-\mathbb{E}_{a\sim\pi_h}[H(Y\mid h,a)].
\]
If $Y\mid a$ is approximately Gaussian with common variance $\sigma^2$ and nearby means $q_a$, then
\[
I(A;Y\mid h)\approx \frac{1}{2\sigma^2}\mathrm{Var}_{a\sim\pi_h}[q_a].
\]
The gradient version replaces $q_a$ with $\psi_h(a)q_a$. Thus value dispersion is a small-noise approximation to reward information, while EPIG-grad is a small-noise approximation to gradient information.

\subsection*{Occupancy correction}
Tree expansion can overrepresent rare prefixes. If a tree allocates many leaves to a prefix reached by only one root chain, a naive tree advantage gives that prefix too much loss mass. Occupancy weighting by $\mu_h$ corrects this. In LLMs, exact occupancy is hard to estimate because token prefixes are rarely revisited exactly; practical estimators include root-chain visitation counts, depth discounting, or segment-level grouping.

\section*{Experiment details}
The control sweep spans thirteen Gym/MuJoCo environments at branch budget 8 with a 32-suffix-per-action reference; EPIG/value-information wins gradient-MSE in 9/13 (all nine dense continuous-control), recovering the reference direction at cosine 0.998--1.000. The frozen GSM8K calibration uses Qwen3-8B over 64 prompts and 3 seeds against a 64-leaf reference. The single-turn math runs use Qwen-class models on GSM8K-hard and MATH-small. The classic-RL ablation comprises a 47-environment flat-GRPO baseline and a seven-environment stochastic-branching $p$-ablation (21 values $\times$ 3 seeds). The loss-mask fix described above and a $k_1\to k_3$ KL-estimator change were applied before the online multi-turn runs.

\section*{Recommended replication protocol for the online test}
For independent replication of the online multi-turn result we recommend: methods flat GRPO, uniform turn tree, EPTree entropy, and EPIG-grad; a Wordle-tuned model with headroom; 16 leaves per prompt, $K_0=4$ root chains, and a fixed token/env-step budget; at least three seeds over the full update budget; and metrics covering reward AUC versus environment interactions, win rate, invalid action rate, KL, entropy, gradient norm, and frozen calibration at checkpoints.


\begin{thebibliography}{8}
\bibitem{deepseekmath} Z. Shao et al., ``DeepSeekMath: Pushing the limits of mathematical reasoning in open language models,'' arXiv:2402.03300, 2024.
\bibitem{treerl} Z. Hou, Z. Hu, Y. Li, R. Lu, J. Tang, and Y. Dong, ``TreeRL: LLM reinforcement learning with on-policy tree search,'' arXiv:2506.11902, 2025.
\bibitem{treepo} Y. Li et al., ``TreePO: Bridging the gap of policy optimization and efficacy and inference efficiency with heuristic tree-based modeling,'' arXiv:2508.17445, 2025.
\bibitem{sutton1999} R. S. Sutton, D. McAllester, S. Singh, and Y. Mansour, ``Policy gradient methods for reinforcement learning with function approximation,'' in \emph{Advances in Neural Information Processing Systems}, 1999.
\bibitem{ppo} J. Schulman, F. Wolski, P. Dhariwal, A. Radford, and O. Klimov, ``Proximal policy optimization algorithms,'' arXiv:1707.06347, 2017.
\bibitem{entropymechanism} G. Cui et al., ``The entropy mechanism of reinforcement learning for reasoning language models,'' arXiv:2505.22617, 2025.
\bibitem{entropyminority} S. Wang et al., ``Beyond the 80/20 rule: High-entropy minority tokens drive effective reinforcement learning for LLM reasoning,'' arXiv:2506.01939, 2025.
\bibitem{gao2023} L. Gao, J. Schulman, and J. Hilton, ``Scaling laws for reward model overoptimization,'' in \emph{Proceedings of the 40th International Conference on Machine Learning}, PMLR 202, 2023, pp. 10835--10866.
\bibitem{epig}
F. Bickford Smith, A. Kirsch, S. Farquhar, Y. Gal,
A. Foster, and T. Rainforth,
``Prediction-oriented Bayesian active learning,''
in \emph{Proceedings of the 26th International Conference
on Artificial Intelligence and Statistics},
Proceedings of Machine Learning Research, vol. 206,
2023, pp. 7331--7348.
\url{https://proceedings.mlr.press/v206/bickfordsmith23a.html}
\end{thebibliography}
\end{document}